# HPSLPred: An Ensemble Multi-label Classifier for Human Protein Subcellular Location Prediction with Imbalanced Source

Shixiang Wan[1], Quan Zou[1]
[1]Tianjin University, Tianjin, China
Email: shixiangwan@gmail.com, zouquan@nclab.net

**Abstract:** Protein subcellular localization prediction is an important and challenging problem. The traditional biology experiments are expensive and time-consuming, so more and more research interests tend to a series of machine learning approaches for predicting protein subcellular location. There are two main difficult problems among the existing state-of-the-art methods. First, most of the existing techniques are designed to deal with the multi-class but not the multi-label classification, which ignores the connection between the multiple labels. In reality, multiple location proteins implicate that there are vital and unique biological significances worthy of special focus, which cannot be ignored. Second, the techniques for handling imbalanced data in multi-label classification problem is significant but less. For solving the two issues, we have developed an ensemble multi-label classifier called HPSLPred which can be applied for the multi-label classification with imbalanced protein source. For the conveniences of users, a user-friendly webserver for HPSLPred was established at *http://server.malab.cn/HPSLPred*.

## 1. Introduction

Identification of protein subcellular localization has a key role in genome annotation, function prediction and vaccine target identification. In briefly, there are two significant aspects. First, the rare and necessary perceptiveness or hints can be discovered from subcellular location information. Normally, the proteins have the exclusive localizations after synthesized in ribosome[1]. If not, the aberrant subcellular location may result in diseases[2]. Second, the interaction between proteins exposes the molecular function mechanism and complex physiological processes[3]. Recognizing the protein functions based on the context subcellular location environment is one of the most important and hottest research issues in proteomics and drug design research[4].

The traditional biology experiments like cell culture, cell separation and protein extraction are costly and time-consuming[5], so more and more research interests tend to a series of machine learning approaches for predicting protein subcellular location. Meanwhile, protein sequences indexed has an immense number. Specifically, there are 553,231 protein sequences now according to the release on 01-Dec-2016 at UniProtKB[6], which is far more than 3,939 protein sequences in 1986. With the rapid increment of protein sequences, there is a strong demand to evolve more accurate and efficient computational approaches based on machine learning.

During the past several decades, a series of approaches surged on subcellular localization[7-14], which have indelible contributions on bioinformatics though these works have their own limitations. Simply, we divide all state-of-the-art works into two types: (1). work on improving precision based more subcellular location sites, which is from only two sites[8] to five sites[9], to twelve sites[10,15] and so on. (2). integrate more different feature methods to make sequences express more comprehensively, such as the amino acid composition method[9,16], the pseudo amino acid composition method[17], the various modes[18-26], and the sequential evolution information[27].

Although it is abundant in subcellular location methods, two main issues need to be ameliorated. First, most methods are based on multi-class prediction line, which implicates that a protein sequence will only be classified in one location site. Actually, a lots of protein sequences exist in multiple sites simultaneously, which has proved by Millar et al[28]. We cannot ignore the influences in separate sites, maybe they act a unique biological role in one site and worthy of particular research[29,30]. Second, the techniques for handling imbalanced data in multi-label classification problem is significant but less.

Imbalanced source may result in terrible recall rate for the fewer samples' label and discrediting the classification results[31]. Based on machine learning technique, we propose a novel and ensemble multi-label classifier HPSLPred for human protein subcellular location prediction with imbalanced source to improve such two shortcomings. Generally, there are four parts to construct a protein subcellular location prediction model[32]. First, data collection and cleaning. Second, feature extraction from protein sequences. Third, blueprint a multi-label classifier for prediction. Forth, build a user-friendly webserver.

In this paper, we propose a novel and ensemble classifier with imbalanced source for human protein subcellular location prediction. The second section of this paper depicts the procedure of constructing original subcellular location source. In this section, after data processes, data cleaning and redundancy eliminating, we acquired 9,895 human protein sequences with the lowest similarity as the test dataset. Next, we introduce the algorithms of protein sequences feature extraction, the method about how to handle the imbalanced source, performance metrics and the novel approach of building multi-label ensemble classifier called HPSLPred, which is based on the compound features and optimal dimension searching. The third section shows the detailed results of HPSLPred compared with 188-dimensional classical features and PseAAC features. After comparing with state-of-the-art methods, our method achieves a higher accuracy, a lower running time and more user-friendly web server. The last section describes the future developmental trend on protein subcellular localization research.

## 2. Material and Methods

### 2.1 Data construction with human protein

Table 1. The first fifty human subcellular location sorted by number

| Ordinal | Subcellular location | Proteins' number | Ordinal | Subcellular location | Proteins' number |
|---|---|---|---|---|---|
| **1** | **Cytoplasm** | **4,410** | 26 | Lysosome membrane | 131 |
| **2** | **Nucleus** | **4,023** | 27 | Postsynaptic cell membrane | 131 |
| **3** | **Cell membrane** | **1,909** | 28 | Nucleus speckle | 125 |
| **4** | **Membrane** | **1,807** | 29 | Nucleoplasm | 111 |
| **5** | **Secreted** | **1,255** | 30 | Mitochondrion matrix | 111 |
| **6** | **Cytoskeleton** | **1,102** | 31 | Secretory vesicle | 105 |
| **7** | **Cell projection** | **755** | 32 | Endosome membrane | 105 |
| **8** | **Endoplasmic reticulum membrane** | **601** | 33 | Spindle | 104 |
| **9** | **Cell junction** | **583** | 34 | Centromere | 101 |
| **10** | **Mitochondrion** | **400** | 35 | Dendrite | 100 |
| 11 | Golgi apparatus | 377 | 36 | Apical cell membrane | 94 |
| 12 | Cytoplasmic vesicle | 373 | 37 | Mitochondrion outer membrane | 93 |
| 13 | Synapse | 341 | 38 | Cytoplasmic vesicle membrane | 91 |
| 14 | Golgi apparatus membrane | 331 | 39 | Axon | 81 |
| 15 | Microtubule organizing center | 319 | 40 | Cilium basal body | 80 |
| 16 | Nucleolus | 315 | 41 | Early endosome | 80 |
| 17 | Centrosome | 314 | 42 | Trans-Golgi network membrane | 80 |
| 18 | Cytosol | 291 | 43 | Nucleus membrane | 80 |
| 19 | Chromosome | 267 | 44 | Late endosome membrane | 79 |
| 20 | Endoplasmic reticulum | 207 | 45 | Early endosome membrane | 74 |
| 21 | Extracellular space | 203 | 46 | Endosome | 74 |
| 22 | Perinuclear region | 185 | 47 | Lysosome | 73 |
| 23 | Mitochondrion inner membrane | 177 | 48 | Melanosome | 70 |
| 24 | Extracellular matrix | 164 | 49 | Postsynaptic density | 68 |
| 25 | Cilium | 141 | 50 | Centriole | 67 |

A series of databases for protein subcellular location, such as UniprotKB[6], LOCATE[33], PSORTdb[34], Arabidopsis Subcellular DB[35], Yeast Subcellular DB[36], Plant-PLoc[37], LOCtarget[38], LOC3D[39], DBSubloc[40], PA-GOSUB[41] and so on. We collected human protein sequences through API provided from the UniprotKB database which is popular and comprehensive. After eliminating repeated samples, the number of original sequences is 11,689, including approximately 200 subcellular location. We statistic and sort each location's number, then pick out the first ten locations as labels of protein sequences. The chosen ten locations respectively are: cytoplasm, nucleus, cell membrane, membrane, secreted, cytoskeleton, cell projection, endoplasmic reticulum membrane, cell junction and mitochondrion. For more convenience expression, we name these locations respectively as: Class1, Class2, Class3, Class4, Class5, Class6, Class7, Class8, Class9 and Class10. The first fifty locations under space limitation and chosen ten locations are shown in Table 1.

Since parts of protein subcellular locations do not belong to anyone of chosen ten locations, such sequences had been removed and 10,613 protein sequences were left. Then setting variant threshold and getting rid of redundance among sequences by CD-HIT[42], we acquired 9,895 samples when the lowest threshold value is 0.7, which is our final test dataset $\mathcal{D}$. Compared with the integrated source[1], the number of protein sequences in our dataset $\mathcal{D}$ exceeds the sample's number in their work. All location numbers under different threshold value are shown in Table 2.

Table2. All location numbers under different threshold value in CD-HIT

| Threshold value | Proteins' number | Class1 | Class2 | Class3 | Class4 | Class5 | Class6 | Class7 | Class8 | Class9 | Class10 |
|---|---|---|---|---|---|---|---|---|---|---|---|
| 0.9 | 10,324 | 3,523 | 3,687 | 1,755 | 1,740 | 1,151 | 838 | 544 | 573 | 467 | 397 |
| 0.8 | 10,132 | 3,459 | 3,626 | 1,693 | 1,716 | 1,113 | 822 | 534 | 564 | 459 | 395 |
| **0.7** | **9,858** | **3,374** | **3,520** | **1,611** | **1,677** | **1,090** | **800** | **521** | **550** | **443** | **392** |

From Table 2, we can be aware of that there are parts of protein sequences belong to multiple labels. Suppose that the number of belonging to multiple i labels is $N_i$ ($i = 1,2,3 \dots 10$), we statistic $N_i$ in Table 3.

Table 3. The number of belonging to multiple i labels

| | Single-label | Multi-label | | | | | | | | |
|---|---|---|---|---|---|---|---|---|---|---|
| | $N_1$ | $N_2$ | $N_3$ | $N_4$ | $N_5$ | $N_6$ | $N_7$ | $N_8$ | $N_9$ | $N_{10}$ |
| Proteins' number | 6,845 | 2,209 | 583 | 158 | 48 | 11 | 4 | 0 | 0 | 0 |

$N_1$ means the sample with belonging to one label, which is the single-label. $N_i$ ($2 \leq i \leq 10$) means the multi-label sample. The number of multi-label samples is about 30.60% of all samples, which cannot be ignored.

In traditional subcellular location works, many ripe methods converted a multi-label job into a multi-class job, relates a sample $S_i$ ($i \in n, n\ is\ the\ total\ number\ of\ all\ samples$) with a single-label $L_j$ ($j \in l, l\ is\ the\ total\ number\ of\ all\ labels$). A single-label dataset $S$ is made up of n samples like $(S_i, L_j)$. It is apparent that multi-class classification will bring forth repeated sequences source when expressing different labels, which is superfluous to classifier. To solve this tissue, many approaches converted a multi-label job into more single-label jobs. All results will be converted back into multi-label statement after single-label classifications. There are a series of state-of-the-art conversion methods, such as Support Vector Machines[43], Naive Bayes[44] and k Nearest Neighbor[45] methods. We will consider more single-label classifiers and multi-label classifiers with our dataset $\mathcal{D}$.

*2.2 Handling imbalanced source*

Given that subcellular location problem is a multi-label problem, many methods transformed this problem into more single-label problems. There are a series of state-of-the-art methods on this issue, and all of them can be summarized as two fundamental methods[46]: the label combination method[47,48] (CM) and the binary relevance method[49-51] (BM).

For CM method, abstract and united (atomic) label is integrated from the original multiple labels. The label $L_j (j \in l)$ will be transformed into $\mathcal{L}$, which is the atomic label corresponding a diverse label subset. Advantage of this methods is that the coaction of all labels can be expressed as rich as possible, but it loses sight of individual influence which is crucial for subcellular location problem.

For BM method, it converses a multi-label issue into multiple one binary issues based each label. Multiple binary classifiers ($C_1, C_2, \dots C_l$) will be trained respectively, each of which predict a binary value linked with each label $L_j \in l$. Our dataset $\mathcal{D}$ can be formulated as:

$$\mathcal{D} = \mathcal{D}_1 \cup \mathcal{D}_2 \cup \mathcal{D}_3 \cup \dots \cup \mathcal{D}_l \ (l\ is\ total\ number\ of\ all\ labels) \quad (1)$$

Disadvantage of this methods is that BM disregards label interactions among every labels when forecasting. But for our dataset, there are no necessary to construct bridges for different subcellular locations since they are independent of each other. In addition, BM maybe generate imbalanced datasets for each label after transforming processes, which can be overcome showed in this paper but only more time complexity. Therefore, BM-based methods will be applied in our approach.

Now, we will describe particular procedure about how to handle the imbalanced source generated by BM methods. The methods of handling imbalanced source are mainly two basic issues: (a). create a new classifier algorithm or modify one or more the existing algorithms for transforming imbalanced presentation. (b). improving data preprocess such as resampling for dispelling imbalanced presentation. Modifying algorithm needs more complex work or more comprehensive source to validate its performance, while improving data preprocesses are more elastic and better in aiming at a specific problem. Besides, there are a series of ensemble approaches, such as altering ensemble classifier algorithm during some learning stages and inserting cost-sensitive learning process, applied for dealing with imbalanced datasets.

Empirically, some works have proved that resampling techniques are valid for equilibrating the class distribution. Moreover, these techniques are standalone with the specific classifier. Resampling techniques can be categorized into three families[31]: (a). Under-Sampling (US). (b). Over-Sampling (OS). (c). Hybrid-Sampling (HS). US method produces a subset from the original dataset to cut down superfluous class samples based majority class samples. OS methods produces a superset from the original dataset to increase majority class samples based superfluous class samples. Hence HS method will integrate US method and OS method and produce an eclectic dataset. Actually, the easiest preprocesses methods from all such three families are random under sampling and random over sampling. The weakness of the former is that all samples cannot be full use of classification, which will not produce an optimal dataset. Additionally, the later maybe result in overfitting because of existing sample copies according to some works[52].

For solving the shortcoming of random over sampling, several methods were designed, such as 'Synthetic Minority Oversampling Technique' (SMOTE[53]), Borderline-SMOTE[54], Adaptive Synthetic Sampling[55], Safe-Level-SMOTE[56] and SPIDER2[57] approaches. However, the same shortcoming of these methods is short of consideration about neighboring instances, which will result in the increment of overlapping between classes. Regarding random under sampling, based data cleaning line, the proposed approaches trend to remit the influence of losing useful data, such as the Wilson's edited nearest neighbor (ENN[58]), the one-sided selection (OSS[59]), Support Vector Machines (SVM[43]) and so on. The main idea about these methods is that finding the superfluous class samples whose number is as same as the majority class samples and close to the decision boundary, combined balanced dataset with the majority class samples. According to figure 1, searching the balanced samples near the decision boundary based support vector machine algorithm is our approach on handling imbalanced source.

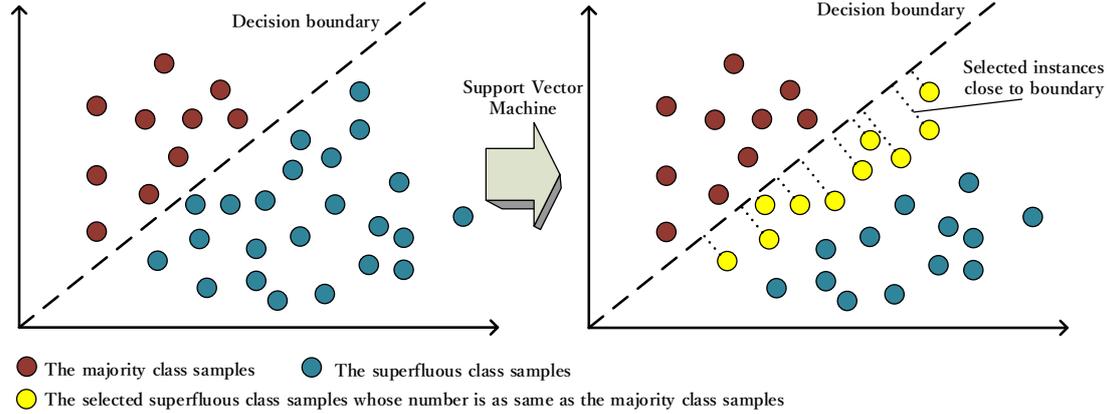

Figure 1. Searching the balanced samples near the decision boundary based SVM algorithm

As a popular supervised learning algorithm, SVM was put forward by Vapnik[60]. SVM can generate a linear decision boundary to tell from all samples[61]. Additionally, the new test samples can be distinguished from classification regulation. If the test samples are not separable, the kernel function is able to map the samples to a high-order feature space until the optimal decision boundary can be ensured. There has been a mature theory and practice foundation on SVM methods, which is more understandable and convincing than loosely heuristic algorithms like some black boxes.

In this study, the LIBSVM package[62] severs as an realization of SVM. The radial basis function (RBF) is taken as the kernel function, which is defined as:

$$\mathcal{K}(x_i, x_j) = exp\left(-\gamma \|x_i - x_j\|^2\right) \tag{2}$$

The values of $\gamma$ and regularization parameter $\mathcal{C}$ in RBF function are optimized on the dataset by cross-validation. For the multi-state classification, the one-versus-one strategy is used.

Up to now, we have particularly described about how to transform a multi-label problem with imbalanced source into multiple problems with balanced source. Suppose that a balanced source for the special label $i$ dealt with SVM is $\mathcal{D}_i^{\mathcal{B}} (i \in l)$, which can be formulated as:

$$\mathcal{D}^{\mathcal{B}} = SVM(\mathcal{D}_1) \cup SVM(\mathcal{D}_2) \cup SVM(\mathcal{D}_3) \ldots \cup SVM(\mathcal{D}_l) \quad (3)$$

$$\mathcal{D}_i^{\mathcal{B}} = SVM(\mathcal{D}_i) \; (l \text{ is total number of all labels}) \quad (4)$$

*2.3 Features for subcellular localization*

In this section, all human protein sequences in dataset $\mathcal{D}$ will be extracted feature vectors based on machine learning. For giving more comprehensive consideration to protein sequences' properties, our approach consider two state-of-the-art feature models: 188-dimension[63] and Pse-in-One[64,65].

188-dimension feature model contains eight types of physical chemical properties, respectively, hydrophobicity, normalized van der Waals volume, polarity, polarizability, charge, surface tension, secondary structure, and solvent accessibility. For the components of 188-dimension, the first 20 dimensions stand for the proportions of the 20 kinds of amino acids, the rest dimensions are made up of eight kinds of more comprehensive physical chemical properties which contain twenty-one attributes. Hence, each protein sequence will bring forth 188 (20 + (21) × 8) numerical values. Pse-in-One feature model contains seven extraction methods, respectively, auto covariance (AC, 22D), cross covariance (CC, 22D) and auto-cross covariance (ACC, 22D) which based on autocorrelation technique; parallel correlation pseudo amino acid composition (PC-PseAAC, 22D), series correlation pseudo amino acid composition (SC-PseAAC, 26D), general parallel correlation pseudo amino acid composition (PC-PseAAC-General, 22D) and general series correlation pseudo amino acid composition (SC-PseAAC-General, 26D) which based on pseudo amino acid composition technique.

*2.4 HPSLPred: a novel and ensemble multi-label classifier*

HPSLPred is an ensemble multi-label classifier with imbalanced source, and we design more comprehensive 350D feature model, automatic optimal dimension searching and ensemble classifiers for it to achieve the best performance.

Based on the dataset transforming method introduced in data construction, the feature model of our approach integrated all seven extraction methods in Pse-in-One and 188-dimension methods, then produce 350D (188D + 5 × 22D + 2 × 26D = 350D) hybrid features model for designing HPSLPred, which expresses protein sequences' properties sufficiently in different viewpoints. Hybrid features model can be seen in some works, and they can achieve more accurate prediction but slower.

However, it is no deny that hybrid features contains some redundant expressions, so we apply an unsupervised dimension reduction method called Max-Relevance-Max-Distance(MRMD[66]) for trimming down dimensionality. MRMD method picks out the target samples with low redundancy and strong relevance from by the large amount of relevance (Max-Relevance) and distance (Max-Distance) rule. Pearson's correlation coefficient is employed for calculating the relevance value, and three kinds of distance functions are adopted to calculate the redundancy value. Actually, the larger relevance, the larger feature's distance and the lower subset's redundancy. Suppose that the labels' number is $\ell$, the optimal condition of MRMD algorithm can be formulated as:

$$\max(MR_i + MD_i) \; (i \in \ell) \quad (5)$$

Sometimes the key point of the specific issues is diverse on estimating the importance between MR and MD, hence adding weights on the original criterion is a wisely strategy. The new criterion can be formulated as:

$$\max(w_r \times MR_i + w_d \times MD_i) \quad 6)$$

where the variables $w_r (0 < w_r \leq 1)$ and $w_d (0 < w_d \leq 1)$ are the weights of MR and MD respectively.

Based on the 350D feature model and MRMD dimension reduction algorithm, automatic optimal dimension searching[67] which is integrated into HPSLPred framework. The whole automatic optimal dimension searching procedure is conceived as some multiple thread processes for the lowest time complexity. Suppose that there is a balanced dataset whose dimension is 350D, the first step, we will run MRMD algorithm to cut down dimension to 10D and compute the performance metrics at such state. Next, elevate 10D to the new dimension from the former, continue to cut down to such new dimension and compute the performance metrics again. Dimension reduction process is designed as eight separate calculating threads which calculates at the same time. In other words, this approach will calculate 80 dimensional span, which makes time complexity far lower. Finally, the first round of the whole procedure named roughly process is accomplished after achieving 340D. Meanwhile, this methods will search a space whose dimensional span is 10D, then the second round named particular process will go on in such span, which is designed as ten separate calculating threads. After the second round, our method will search the best dimension according to the optimal performance metrics. The whole procedure is called the two layer optimal dimension searching, which has proved as a not bad method.

In many state-of-the-art works of machine learning tools, Mulan[68], WEKA[69] and scikit-learn[70] are excellent up to now. In this study, we reconstruct an ensemble multi-label classifier HPSLPred with the help of such three software tools. Advantage of Mulan is that it can classify multi-label dataset directly and has an intimate connection with WEKA, which is the main character we will compare with. Advantage of WEKA is that it integrates fully machine learning algorithms including about classification, cluster, and association, but that multi-label dataset cannot be handled directly is the main shortcoming. Advantage of scikit-learn is that it has a faster running-speed, more flexible interface for developing, even modify the core parts of one algorithm, while it cannot provide friendly user interface, and solution on multi-label dataset is shorted as well. However, after joining our strategies about handling multi-label imbalanced source, scikit-learn's shortcoming has been overcome, and we named such novel approach as HPSLPred based its proud running-speed and better performance. By contrast, (a). Mulan will classify the original multi-label source, and referred classifiers respectively are BRkNN, HOMER, MLkNN, IBLR_ML and DMLkNN. (b). WEKA will classify the multiple single-label balanced dataset transformed from the original multi-label source according to the above, and referred classifiers respectively are IBk, Random Forest and J48. (c). HPSLPred will classify the dataset as same as WEKA's dataset and add the two layer optimal dimension searching approach, and referred classifiers in scikit-learn respectively are Naive Bayes, Logistic Regression, SGD, Decision Tree, Nearest Neighbors, Extra Trees, Random Forest, LinearSVC, Bagging, AdaBoost, Gradient Boosting and LibSVM. Besides, the maximum precision of all such twelve classifiers on one label will be considered as this label's precision. More comparison information will be showed in next section.

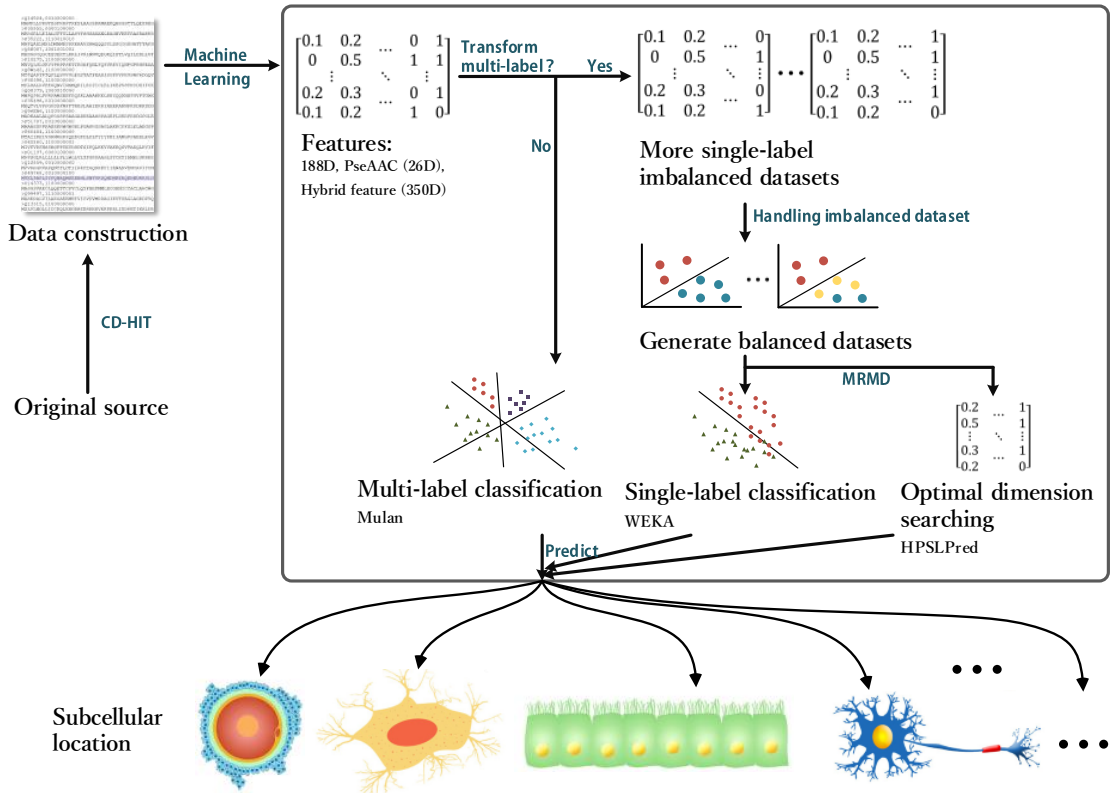

Figure 2. The work flow of our whole approach

*2.5 Performance metrics*

There are several performance metrics on multi-label classification problem, such as Hamming loss, Coverage, OneError, IsError, Ranking loss and Average precision (AP). The metric that the most relevant to multi-label classification system performance is AP value, which ranges 0 to 1. AP metric can be formulated as:

$$AP(f) = \frac{1}{N}\sum_{i=1}^{N}\frac{1}{S_i}\sum_{\xi \in y_i}\frac{|L_i|}{predict_f(x_i, \xi)}$$

$$L = \{\xi' | predict_f(x_i, \xi') \leq predict_f(x_i, \xi), \xi' \in y_i\}$$

Where $N$ is all samples' number; $S_i$ is the number of the samples with label $y_i$; $predict_f(x_i, \xi)$ is the precision on sample $x_i$ with label $\xi$. Average precision will be applied for a mainly standard in a series of comparison experiments.

## 3. Results

Based a series of detailed descriptions in above sections, we will validate whether our approach is better or not. First of all, with 188D feature model, our approach will test the performance of Mulan, WEKA and HPSLPred based scikit-learn. The precision of each label or each subcellular location will be displayed clearly. Similarity, we also show the performance about PseAAC-26D feature model and 350D hybrid feature model, which includes an extra presentation on the two layer optimal dimension searching. Additionally, we modestly compared our approach with some state-of-the-art methods to increase persuasion. Except such comprehensive experiments, we offer all source code in public and a user-friendly webserver to researchers.

### *3.1 Comparative experiments based on 188D classical features*

Table 4. Results of comparative experiments based on 188D classical features

|         | IBk    | RandomForest | J48    | BRkNN  | HOMER  | MLkNN  | IBLR_ML | DMLkNN | HPSLPred |
|---------|--------|--------------|--------|--------|--------|--------|---------|--------|----------|
| class1  | 66.95% | 68.01%       | 64.06% | 60.13% | 47.17% | 68.83% | 69.96%  | 68.72% | 74.33%   |
| class2  | 70.57% | 71.65%       | 68.15% | 70.64% | 52.11% | 72.16% | 71.89%  | 72.73% | 77.19%   |
| class3  | 63.38% | 65.80%       | 60.58% | 65.05% | 46.43% | 66.41% | 64.25%  | 67.15% | 70.71%   |
| class4  | 63.57% | 65.30%       | 62.34% | 62.40% | 45.57% | 66.19% | 66.15%  | 64.50% | 72.05%   |
| class5  | 70.92% | 73.21%       | 68.67% | 72.94% | 55.57% | 73.36% | 73.94%  | 75.03% | 78.53%   |
| class6  | 67.06% | 67.81%       | 61.88% | 69.12% | 48.12% | 69.10% | 70.37%  | 71.22% | 73.04%   |
| class7  | 58.45% | 62.67%       | 61.04% | 59.50% | 40.47% | 63.49% | 62.21%  | 61.59% | 66.24%   |
| class8  | 69.82% | 72.64%       | 64.55% | 66.55% | 50.22% | 68.68% | 72.32%  | 68.64% | 72.84%   |
| class9  | 63.66% | 66.14%       | 59.71% | 63.09% | 39.80% | 62.25% | 63.93%  | 65.19% | 64.88%   |
| class10 | 65.18% | 68.75%       | 67.09% | 63.65% | 51.64% | 68.53% | 71.29%  | 65.74% | 74.65%   |
| AP      | 65.95% | 68.20%       | 63.81% | 66.34% | 47.71% | 67.90% | 68.63%  | 68.05% | 72.45%   |

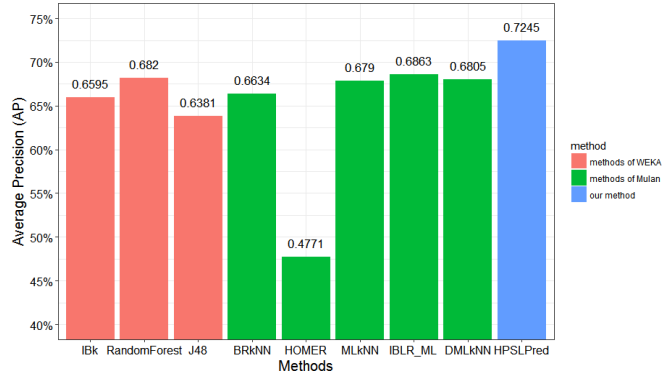

Figure 3. AP values of each classifier with 188D source.

The 188D dimension feature model takes full consideration of the proportions of amino acids and eight types of physical chemical properties, which has proved as an efficient feature model. For our original 188D multi-label source $\mathcal{D}_{188D}$, three experiments will be arranged as following:

Experiment (a). Based on BRkNN, HOMER, MLkNN, IBLR_ML and DMLkNN, Mulan classifies the $\mathcal{D}_{188D}$ directly by ten folds crossing validation.

Experiment (b). As the way of we described, firstly, multi-label source should be processed into more single-label datasets through BM method, which including enough positive samples and negative samples. Then balanced dataset will be extracted by SVM decision boundary from each single-label imbalanced dataset. All balanced datasets is named $\mathcal{D}_{188D}^{B}$. Based on IBk, Random Forest and J48, WEKA classifies the $\mathcal{D}_{188D}^{B}$ directly by ten folds crossing validation.

Experiment (c). Based on Naive Bayes, Logistic Regression, SGD, Decision Tree, Nearest Neighbors, Extra Trees, Random Forest, LinearSVC, Bagging, AdaBoost, Gradient Boosting and LibSVM, HPSLPred classifies the $\mathcal{D}_{188D}^{B}$ directly by ten folds crossing validation. HPSLPred selects the maximum precision of all twelve classifiers as such label's precision through grid parameters searching, then the mean value of all label's precision is average precision.

All experiments' results are displayed in Table 4 and Figure 3. The first conclusion is that balanced datasets have a better performance than multi-label dataset. IBLR_ML achieves the highest precision 68.63% in five multi-label classifiers belonging to Mulan. For three WEKA classifiers, RandomForest achieves 68.20% precision, which is much better than others. HPSLPred makes full use of the ensemble classifiers' advantage and reaches 72.45% precision that exceeds Mulan and WEKA apparently. Each label's precision in HPSLPred is higher than others' classifier, which is not an occasional result. Hence it should attribute to all various classifiers which explains implications of classification in different viewpoints. In order to validate availability based more feature model comprehensively, our experiments will go on.

### 3.2 Comparative experiments based on PseAAC-26D classical features

Table 5. Results of comparative experiments based on PseAAC-26D classical features

|  | IBk | RandomForest | J48 | BRkNN | HOMER | MLkNN | IBLR_ML | DMLkNN | HPSLPred |
|---|---|---|---|---|---|---|---|---|---|
| class1 | 65.55% | 68.79% | 67.65% | 65.55% | 51.70% | 67.76% | 71.30% | 67.46% | 77.02% |
| class2 | 69.18% | 70.97% | 67.26% | 70.49% | 55.59% | 71.45% | 73.86% | 71.10% | 77.46% |
| class3 | 61.11% | 65.21% | 61.45% | 64.81% | 49.14% | 65.32% | 66.75% | 65.25% | 70.65% |
| class4 | 62.67% | 65.59% | 65.83% | 63.95% | 49.78% | 66.31% | 66.27% | 64.75% | 75.08% |
| class5 | 70.55% | 73.58% | 69.17% | 73.95% | 59.17% | 71.87% | 70.64% | 72.66% | 78.80% |
| class6 | 65.88% | 67.63% | 65.13% | 66.50% | 50.81% | 66.94% | 74.99% | 67.27% | 74.69% |
| class7 | 56.53% | 57.68% | 57.10% | 58.85% | 45.52% | 61.31% | 61.98% | 62.12% | 66.26% |
| class8 | 63.18% | 68.27% | 65.82% | 68.60% | 54.80% | 68.25% | 68.81% | 72.09% | 77.58% |
| class9 | 59.71% | 60.84% | 58.01% | 58.18% | 45.81% | 63.41% | 59.67% | 65.59% | 67.32% |
| class10 | 64.92% | 68.24% | 62.88% | 70.02% | 53.77% | 72.38% | 68.45% | 68.20% | 77.20% |
| AP | 63.93% | 66.68% | 64.03% | 66.09% | 51.61% | 67.50% | 68.27% | 67.65% | 74.21% |

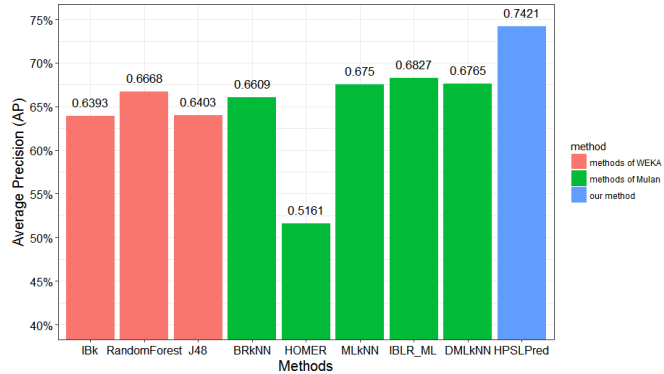

Figure 4. AP values of each classifier with 188D source.

In this subsection, we will show all experiments about PseAAC-26D features model. PseAAC is one of protein sequences features extraction methods based pseudo amino acid composition in Pse-in-One, which has achieved very good results in many previous works. For our original PseAAC-26D multi-label source $\mathcal{D}_{26D}$, three experiments are as following:

Experiment (d). Mulan will sort the $\mathcal{D}_{26D}$ based a series of multi-label classifiers.

Experiment (e). Suppose that $\mathcal{D}_{26D}^{B}$ is all single-label balanced datasets transformed by $\mathcal{D}_{26D}$, WEKA will itemize the $\mathcal{D}_{26D}^{B}$ based a series of single-label classifiers.

Experiment (f). HPSLPred will classify the $\mathcal{D}_{26D}^{B}$ based scikit-learn single-label classifiers.

All experiments' results are displayed in Table 5 and Figure 4. From the information given, it can be concluded that HPSLPred achieves the best AP as same as the experiments showed in last subsection. However, each classifier's AP under PseAAC feature model cannot lump together. There are much difference between WEKA and Mulan. In the comparative experiments in this round, Mulan with multi-label source is still better than Weka with single-label balanced datasets, which fully affirms the previous excellent works. For three classifiers of WEKA, only J48 is better but not the highest one. For five classifiers of Mulan, only HOMER is better but it is still the lowest one. The main shortcoming of traditional single-label classifiers is that they cannot deal with several feature expressions with disparate dataset dynamically. But the ensemble classifier based scikit-learn grid parameters searching, HPSLPred, overcomes this weakness and achieves the highest AP 74.21%, which is about 6% higher than the second best algorithm.

### *3.3 Comparative experiments based on hybrid 350D features and dimensional searching*

Table 6. Results of comparative experiments based on hybrid 350D features

|        | IBk    | RandomForest | J48    | BRkNN  | HOMER  | MLkNN  | IBLR_ML | DMLkNN | HPSLPred |
|--------|--------|--------------|--------|--------|--------|--------|---------|--------|----------|
| class1 | 66.76% | 69.35%       | 64.83% | 67.50% | 47.86% | 69.62% | 69.78%  | 69.47% | 77.67%   |
| class2 | 71.70% | 72.14%       | 67.95% | 71.59% | 52.80% | 72.95% | 73.47%  | 73.48% | 79.77%   |
| class3 | 65.74% | 65.67%       | 61.45% | 64.02% | 47.12% | 67.20% | 67.34%  | 67.90% | 71.38%   |
| class4 | 64.19% | 68.40%       | 63.57% | 65.78% | 46.26% | 66.98% | 68.33%  | 65.25% | 74.44%   |
| class5 | 73.07% | 75.37%       | 68.99% | 72.10% | 56.26% | 74.15% | 73.89%  | 75.78% | 81.56%   |
| class6 | 67.38% | 68.50%       | 63.25% | 65.31% | 48.81% | 69.89% | 68.96%  | 71.97% | 76.73%   |
| class7 | 58.83% | 60.36%       | 59.98% | 64.47% | 41.16% | 64.28% | 63.33%  | 62.34% | 66.00%   |
| class8 | 67.82% | 74.18%       | 65.45% | 67.98% | 50.91% | 69.47% | 70.27%  | 69.39% | 75.47%   |
| class9 | 62.19% | 61.06%       | 60.05% | 63.14% | 40.49% | 63.04% | 65.43%  | 65.94% | 66.08%   |
| class10| 65.56% | 73.21%       | 68.75% | 70.52% | 52.33% | 69.32% | 74.40%  | 66.49% | 76.46%   |
| AP     | 66.32% | 68.83%       | 64.43% | 67.24% | 48.40% | 68.69% | 69.52%  | 68.80% | 74.56%   |

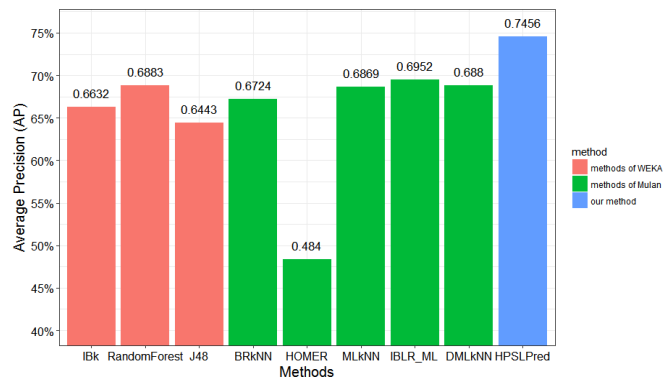

Figure 5. AP values of each classifier with 188D source.

According to the matter in the second section, Pse-in-One contains seven feature extraction methods, and PseAAC is one of all methods. Hybrid 350D feature model is generated from all Pse-in-One methods and 188D methods. Based such hybrid feature model, we will display a series of experiments which is similar with the above. For our original 350D multi-label source $\mathcal{D}_{350D}$, three experiments are as following:

Experiment (g). Mulan will classify the $\mathcal{D}_{350}$ based on five multi-label classifiers.

Experiment (h). Suppose that $\mathcal{D}^B_{350D}$ is all single-label balanced datasets transformed by $\mathcal{D}_{350D}$, WEKA will itemize the $\mathcal{D}^B_{350D}$ based three single-label classifiers.

Experiment (i). HPSLPred will classify the $\mathcal{D}^B_{350D}$ based ensemble scikit-learn single-label classifiers.

All experiments' results are displayed in Table 6 and Figure 5. From the viewpoint of AP performance, the results of 350D feature model improves better than 188D and PseAAC. It illustrates that the degree of coincidence on expressing features has some imparity distributions, and they has themselves' contribution on hybrid 350D model. From the viewpoint of each class's precision, they have some different degrees of improvement, which proves the overall improvement of AP is not an accident but an inner regularity. But it is undeniable that there must be a coincidence among the interpretation. For example, the performance matric of HPSLPred is not as high as other classifiers, so maybe some redundancy exists in it and our approach need to add the two layer optimal dimension searching method to reach the best. As described in last section, for the first round, algorithm will cut down dimension to 10D and compute the performance metrics meanwhile. Scan all dimension by stepping 10D and finish this round. For the second round, searching the each dimension in the highest span in the first round. For our experiments, this span is from 20D to 30D, the optimal AP will bring forth from there. By the way, all procedures we mentioned are integrated into the novel HPSLPred classifier. Figure 6 shows the best AP 75.89%, which is the whole highest average precision we consider about.

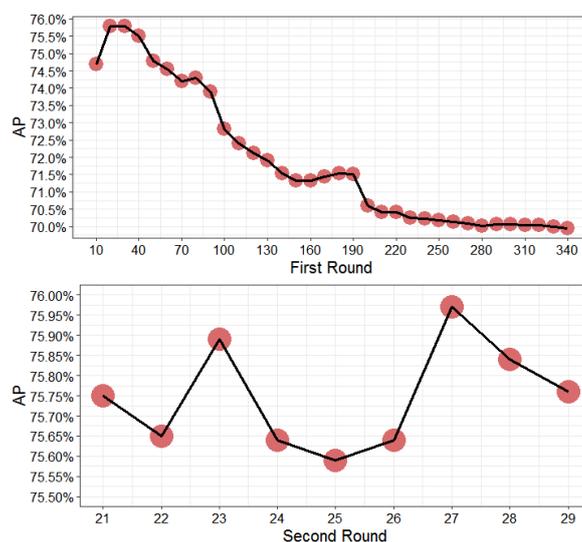

Figure 6. The two layer optimal dimension searching.

### 3.4 Comparison with state-of-the-art methods

Table7. Accuracy comparison with state-of-the-art methods.

| Method | Average Precision |
|---|---|
| Hum-mPLoc 2.0 | 67.95% |
| mGOF-Loc | 65.71% |
| IMMMLGP | 68.76% |
| Xiaotong Guo's approach | 69.52% |
| **HPSLPred** | **75.89%** |

Protein subcellular location research is always one of the hottest issues in cell biology and bioinformatics. Therefore, more and more excellent approaches based machine learning are merged now, such as Hum-mPLoc 2.0[71], mGOF-Loc[72], IMMMLGP[72], and Xiaotong Guo's approach[1]. The first two methods are based on single-label source, while the last two methods are based on multi-label source. For Hum-mPLoc 2.0 and mGOF-Loc methods, we will transform the original multi-label source into more single-label datasets for them. For IMMMLGP and Xiaotong Guo's approach methods, the original source is well. Surprisingly, HPSLPred still achieves the best AP 75.89%, comparatively, Xiaotong Guo's approach is as same as Mulan directly, but its performance is proud. The classical methods such as Hum-mPLoc 2.0, mGOF-Loc and IMMMLGP have not bad performance. Additionally, it can be concluded that the ensemble multi-label classifier has a brighter development prospect than multi-class classifiers in the future.

### 4. Discussion

In this section, a series of experiments showed above will be compared with each other, and we will analysis the merit and defect of each method conclusively. From three experiments (a), (b), and (c) based 188D feature model, we sum up three vital conclusions: (1). HPSLPred classifier integrates twelve kinds of basic classifiers and adds grid parameters searching, which makes it reach the highest AP value. (2). Mulan based multi-label classifiers is better than WEKA based classical single-label classifiers, which proves that Mulan is an excellent work. (3). each label's precision is correspond to its AP value, which eliminate the coincidence of results. From three experiments (d), (e), and (f) based PseAAC-26D feature model, we sum up a new conclusion: there are no apparent relation between PseAAC's results and 188's results, so new expression exists in them. Finally, from three experiments (g), (h), and (i) based hybrid 350D feature model, we sum up a crucial conclusion: hybrid 350D feature model contains redundant expressions, which needs feature selection. After that, we acquire the best AP. The method or classifier based hybrid feature model for protein subcellular location is named HPSLPred. HPSLPred has a convincing performance, especially after a series of comparison with state-of-the-art methods.

## 5. Conclusion

In this study, we generate hybrid 350D feature based 188D feature and all methods of Pse-in-One, transform the original multi-label imbalanced source into multiple single-label imbalanced datasets based BM method, transform single-label imbalanced datasets into single-label balanced datasets based SVM decision boundary, integrate twelve basic classifiers based scikit-learn and match optimal parameters by grid searching, search the optimal dimension by MRMD and two round dimension reduction, finally achieve the highest AP value. Moreover, we utilize multi-threaded technology for accelerating multiple processes in dimensionality reduction. This is our novel and ensemble classifier HPSLPred with imbalanced source. For the conveniences of users, a user-friendly webserver for HPSLPred was established at *http://server.malab.cn/HPSLPred*.

In future work, we want to develop a more comprehensive protein database and keep updated, providing researchers with high-quality research object. Additionally, multi-threaded technology and GPU parallel technology greatly reduces the program's time complexity. Meanwhile, our algorithm strategy should keep pace with the times, make full use of powerful hardware performance and create more efficient multi-label classifiers for improving the precision of subcellular localization problem.

## Acknowledgements

The work was supported by the Natural Science Foundation of China (No. 61370010) and the Natural Science Foundation of Fujian Province of China (No.2014J01253).